\documentclass[runningheads]{llncs}

\usepackage{graphicx}
\usepackage{subcaption}
\usepackage{url}
\usepackage{hyperref}
\usepackage{orcidlink}
\usepackage{amsmath}

\title{Personalized Socially Assistive Robots \\  With End-to-End Speech-Language Models \\ For Well-Being Support }

\author{Mengxue Fu\and
Zhonghao Shi\and
Minyu Huang \and
Siqi Liu \and
Mina Kian \and
Yirui Song \and 
Maja J. Matari\'c}

\authorrunning{Fu et al.}

\institute{Department of Computer Science, University of Southern California, Los Angeles, CA, USA\\
\email{\{mishafu, zhonghas, minyuhua, liusiqi, kian, yfsong, mataric\}@usc.edu}}

\begin{document}

\maketitle

\begin{abstract}
Socially assistive robots (SARs) have shown great potential for supplementing well-being support. However, prior studies have found that existing dialogue pipelines for SARs remain limited in real-time latency, back-channeling, and personalized speech dialogue. Toward addressing these limitations, we propose using integrated end-to-end speech-language models (SLMs) with SARs. This work 1) evaluated the usability of an SLM-enabled SAR dialogue system through a small user study, and 2) identified remaining limitations through study user feedback to inform future improvements. We conducted a small within-participant user study with university students (N = 11) whose results showed that participants perceived an SLM-enabled SAR system as capable of providing empathetic feedback, natural turn-taking, back-channeling, and adaptive responses. We also found that participants reported the robot’s nonverbal behaviors as lacking variability and synchronization with conversation, and the SLM’s verbal feedback as generic and repetitive. These findings highlighted the need for real-time robot movement synchronized with conversation, improved prompting or fine-tuning to generate outputs better aligned with mental health practices, and more expressive, adaptive vocal generation.




\end{abstract}


\section{Introduction}

\begin{figure}[h]
    \centering
    \includegraphics[width=0.8\linewidth]{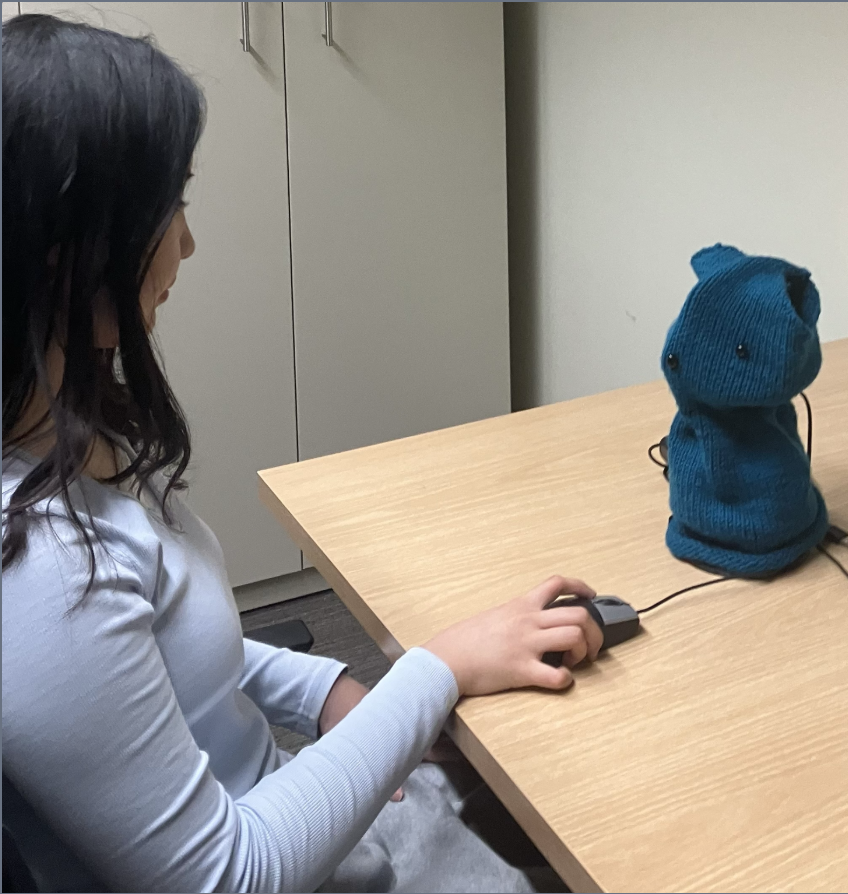} 
    \caption{\textbf{Experiment setup}. The participant interacts with our SLM-enabled SAR system by pressing and holding the mouse to speak, then releasing it to yield the turn to the robot. See Fig.~\ref{fig:workflow} for an overview of the interaction flow.}
    \label{fig:setup}
\end{figure}

Well-being is a fundamental aspect of human health, encompassing emotional, psychological, and social dimensions. In recent years, concerns about well-being have grown in the United States, particularly among young adults, including college students, as mental health issues such as stress, anxiety, and depression continue to rise~\cite{liu2019prevalence}. Despite the increasing need for well-being support, access to mental health care remains limited, particularly for individuals without adequate insurance coverage ~\cite{rowan2013access}. This disparity highlights the need for more accessible and cost-effective well-being support.

A large body of prior work has shown that socially assistive robots (SARs) offer a promising approach to supplement the efforts of mental health professionals by providing accessible and interactive individualized support. Prior research in human-robot interaction (HRI) has demonstrated the effectiveness of SARs in fostering social engagement, emotional support, and positive behavior change~\cite{scoglio2019use,kian2024llm}. However, as shown in Fig.~\ref{fig:mainfigure}, existing SAR systems largely relied on a cascaded dialogue pipeline that connects multiple machine learning (ML) models, including speech-to-text (STT), a dialogue management system, and text-to-speech (TTS)~\cite{spitale2025vita,scheutz2011toward}. This pipeline introduces response latency in robot's verbal speech and non-verbal back-channeling, leading to unnatural turn-taking between the robot and the user. In addition, the pipeline approach made it difficult to personalize verbal feedback and vocal tone in speech output in response to the conversational and emotional context. These limitations contributed to lowered user-perceived levels of empathy, hindering engagement and rapport between the robot and the user, and reducing the effectiveness of SARs for well-being support~\cite{axelsson2024robots}.

To address these limitations, in this work we replace the existing cascaded dialogue approach with a unified end-to-end speech-language model (SLM).  In the validation work we describe, the SLM was GPT-4o-realtime~\cite{gpt4oreal}, which is pre-trained for real-time speech conversation, directly tokenizing audio input and synthesizing audio output without intermediate text representation. This architecture enables low response latency and context-aware speech generation~\cite{gpt4oreal}. Compared to SARs that rely on scripted or cascaded speech processing, which often introduces substantial latency, integration of an end-to-end SLM has the potential to support real-time, adaptive, and expressive conversations that more closely resemble natural human interactions in well-being support settings.
    
To the best of the authors’ knowledge, this is the first attempt to integrate an end-to-end SLM into a SAR and evaluate it in a well-being support context.  This context is challenging because it requires the robot to respond in real time while conveying empathy to ensure effectiveness. For evaluation, we conducted a small N=11 within-subjects study in which the robot guided each participant through a 40-minute gratitude-based exercise for well-being support, followed by a semi-structured participatory design interview. Our quantitative and qualitative results showed that participants perceived the SLM-enabled SAR system as capable of providing empathetic feedback, natural turn-taking, back-channeling, and adaptive responses. Despite these promising findings, our results also showed that the robot’s nonverbal behaviors still lacked variability and synchronization with the speech output from the SLM. Additionally, although the SLM was capable of real-time conversation, its verbal feedback was perceived as generic and repetitive, limiting its ability to build the personal and emotional rapport essential for effective well-being support. These insights suggest that future work is needed to enable real-time robot movement generation synchronized with conversation, develop more sophisticated prompting frameworks and fine-tuning methods to align SLM outputs with evidence-based practices from mental health experts, and to improve SLMs to produce more expressive and adaptive vocal tones in their speech output.


The main contributions of this work are: 1) proposing the use of end-to-end speech-language models (SLMs) for socially assistive robots (SARs) and validating usability through a small user study focused on well-being support; and 2) identifying existing limitations and engaging users in co-design to inform future improvements of SLM-enabled SARs.


\section{Related Work}

\subsection{SARs for Well-Being Support}

A large body of prior work has validated the potential of socially assistive robots (SARs) to supplement the efforts of mental health professionals in providing more accessible well-being support~\cite{bodala2021teleoperated}. Studies have also demonstrated the effectiveness of SARs in promoting psychological well-being and facilitating behavioral change~\cite{scoglio2019use}. For example, Kidd et al.~\cite{kidd2008robots} developed robotic coaches aimed at supporting positive behavioral change over a 4–6 week period, specifically targeting dietary habits. Their findings highlighted the potential of SARs to support sustained behavioral modifications.

Prior work has further explored the use of SARs in well-being and mindfulness interventions. For example, Jeong et al.\cite{jeong2020robotic} developed a well-being coach for college students using the robot Jibo and found significant improvements in participants’ overall psychological well-being. Similarly, Bodala et al.\cite{bodala2021teleoperated} compared a robotic mindfulness coach to a human one, noting that while the human coach was rated higher, both interventions led to positive outcomes. Moreover, work by Spitale et al.\cite{spitale2025vita} explored personalization in SAR-based well-being coaching, demonstrating that adaptive interactions over a four-week longitudinal study improved engagement and well-being outcomes. Additionally, Kian et al.\cite{kian2024llm} demonstrated that a large language model (LLM)-powered SAR delivering daily cognitive behavioral therapy (CBT)-based at-home exercises achieved significant reductions in psychological distress and short-term anxiety among university students, outperforming chatbot and worksheet baselines. Their findings validated the potential of LLM-enabled SARs to improve adherence and therapeutic outcomes in mental health interventions.

\subsection{Dialogue Management Pipelines for SAR: Cascaded vs. End-to-End Speech}

\begin{figure*}[t]
    \centering
    \begin{subfigure}{1.0\linewidth}
        \includegraphics[width=\linewidth]{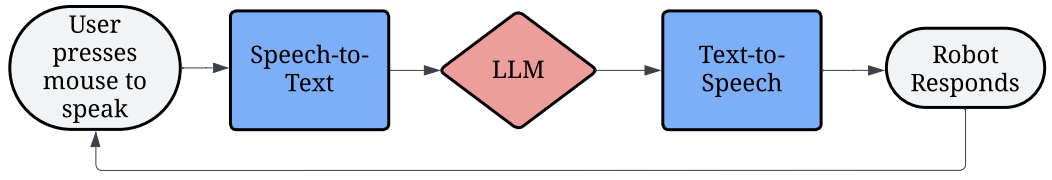}
        \caption{Existing Cascaded Pipeline}
        \label{fig:figure1a}
    \end{subfigure}
    
    \vspace{0.5cm} 

    \begin{subfigure}{1.0\linewidth}
        \includegraphics[width=\linewidth]{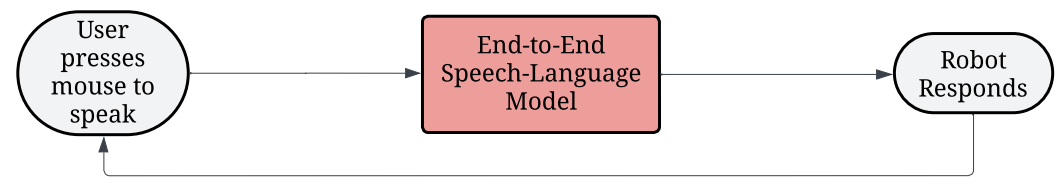}
        \caption{End-to-End SLM}
        \label{fig:figure1b}
    \end{subfigure}

    \caption{\textbf{Comparison between the cascaded dialogue pipeline and an end-to-end SLM for SARs.} SLM enabled real-time inference and speech generation, offering great potential for more natural dialogue in HRI.}
    \label{fig:mainfigure}
\end{figure*}

According to a published survey on dialogue management in HRI~\cite{reimann2024survey}, the dialogue management systems used in HRI often followed a cascaded pipeline, as shown in Fig.~\ref{fig:figure1a}, consisting of the following steps: 1) the user’s speech was first processed by the spoken language understanding module often using a speech-to-text (STT) model to transcribe spoken input into natural language transcription; 2) the transcription was then fed into a dialogue management module, which may be a language model (LM), to generate the robot’s textual response; and 3) the textual response was then passed to the speech generation module, which often used a text-to-speech (TTS) model to produce the corresponding robot speech.

Using this cascaded pipeline, Xu et al. \cite{xu2025exploring} developed the conversational capabilities of a SAR for diary studies, where each step is handled by a separate machine learning model. Similarly, prior work on dialogue systems in HRI by Scheutz et al.\cite{scheutz2011toward} has explored the use of this cascaded dialogue management pipeline to enable real-time human-robot interactions. Although this pipeline successfully supported conversational interaction between humans and robots, they found that the inference time introduced by each module contributed to latency and negatively affected the usability of the conversational system. Prior work on SARs for well-being by Spitale et al.~\cite{spitale2025vita} integrated LLM-based dialogue generation for adaptive well-being coaching, replacing rule-based, predefined dialogue trees. Their system, VITA, demonstrates the potential of multi-modal SAR coaching; however, it also highlights similar limitations, including unnatural turn-taking caused by response latency. These issues hindered the robot’s perceived empathy and its ability to establish rapport with human users.

Similar findings were discussed in the design study on SARs for well-being support by Axelsson et al.~\cite{axelsson2024robots}. The authors found that the limitations of existing cascaded dialogue management pipelines can be summarized by the following key points~\cite{axelsson2024robots}:

\begin{itemize}
\item \textbf{Response latency and unnatural turn-taking:} Awkward pauses or interruptions caused by latency disrupt the flow of conversation and make interactions feel unnatural.
\item \textbf{Inadequate back-channeling:} Also due to latency, robots that do not exhibit synchronized back-channeling behaviors are perceived as inattentive, diminishing the user’s sense of being heard and reducing perceived empathy.
\item \textbf{Lack of personalization:} When dialogue management system relies solely on rule-based, pre-defined dialogue trees instead of a language model, the robot’s responses are not adapted or personalized to the user’s input or conversational context.
\item \textbf{Lack of voice expressiveness:} The robot’s voice lacks emotional expressiveness and is not matched with the contextual information or emotional tone of the speech.
\end{itemize}

Advances in end-to-end SLMs~\cite{achiam2023gpt,fang2024llama} trained directly on speech input and output made these models candidates for effectively eliminating the cascaded structure of traditional dialogue pipelines. This substitution enabled real-time inference and speech generation, offering great potential for more natural dialogue in HRI. To the best of our knowledge, no existing work has yet utilized end-to-end SLM for SAR, particularly in the context of well-being support. Fig.~\ref{fig:mainfigure} illustrates the differences between the existing cascaded pipeline and our proposed end-to-end SLMs.

\section{Methods}





\subsection{Integration of End-to-End SLMs with SARs}

\begin{figure*}[h]
    \centering
    \includegraphics[width=\textwidth]{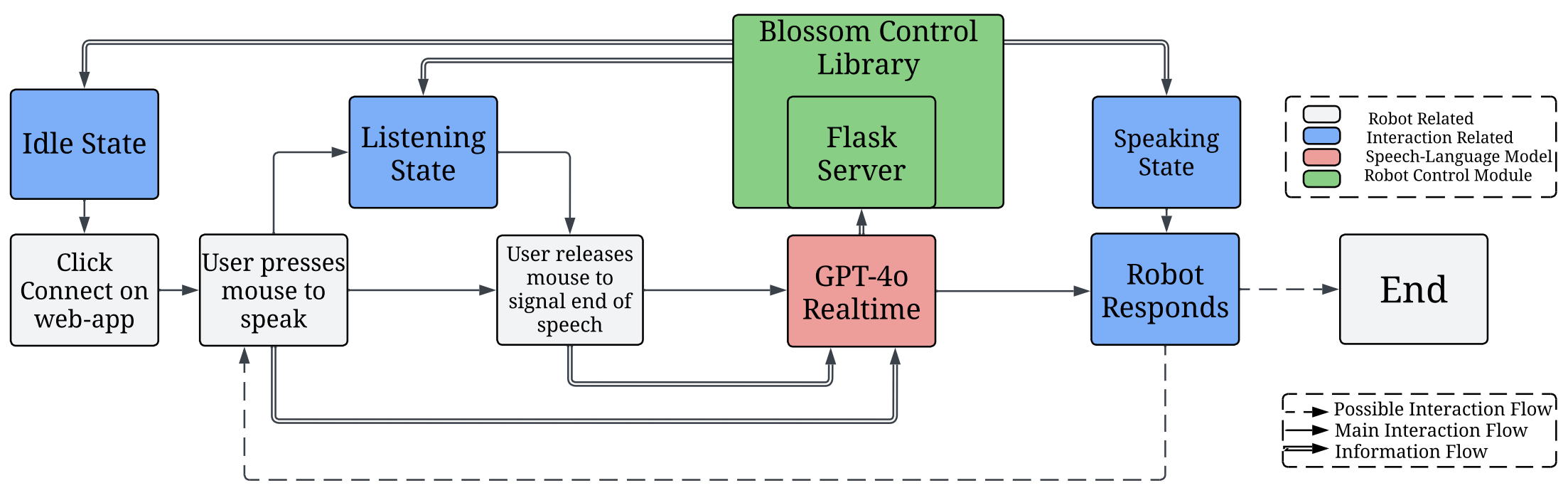}
    \caption{Overview of the integration of an end-to-end SLM with \textit{Blossom}. \textit{Blossom} was connected to a SLM, GPT-4o-realtime, through the real-time API provided by OpenAI. A Python Flask server was used to enable synchronized back-channeling movements of the robot.}
    \label{fig:workflow}
\end{figure*}

In this work, we used a modified version of the \textit{Blossom} robot~\cite{shi2024build}, a handcrafted, open-source robot made of 3D-printed parts and wool, and capable of expressive movements \cite{suguitan2019blossom}. We chose GPT-4o-realtime, accessed through OpenAI’s real-time API~\cite{gpt4oreal,openai_realtimeapi}, because, to the best of our knowledge, it was the only commercially available speech-language model with real-time capabilities at the time of our system setup and development, and there were no comparable proprietary or open-source alternatives. As shown in Fig.~\ref{fig:workflow}, \textit{Blossom} was connected to a web application interfacing with OpenAI’s GPT-4o-realtime via its API. We implemented three non-verbal robot movements--idle, speaking, and listening--that corresponded to different states of the interaction:


\begin{itemize}
\item \textbf{Idle State} – The robot expands and contracts slightly to simulate breathing.
\item \textbf{Speaking State} – The robot shakes its head sideways while generating speech.
\item \textbf{Listening State} – The robot nods while waiting for the participant’s response, indicating active listening.
\end{itemize}

As shown in Fig.~\ref{fig:workflow}, movement-related modules are shown in blue, interaction components are shown in grey, the speech-language model is shown in red, and the robot control module is shown in green.

During the interaction, the \textit{Blossom} robot transitioned between three movement states (idle, listening, and speaking; highlighted in blue) based on the context of the conversation. When no one was speaking, the robot entered the idle state and performed subtle breathing motions to indicate it was powered on and awaiting input. When the participant pressed and held the mouse to speak, the Flask server (highlighted in green) received a signal from the SLM (GPT-4o-realtime, shown in red) and the robot entered the listening state, during which it nodded to convey active listening. When the participant released the mouse, signaling the end of their turn, the robot entered the speaking state and generated speech while performing side-to-side head movements. These synchronized non-verbal behaviors aligned with turn-taking patterns and provided real-time back-channeling cues, aiming to enhance the robot’s perceived attentiveness and ability to build rapport. Full documentation and code are available at: \url{https://github.com/interaction-lab/RealTime_SAR_Well-being_Study}\footnote{We developed our code based on the OpenAI real-time API beta repo: \url{https://github.com/openai/openai-realtime-api-beta}}.

\subsection{Research Hypotheses}

The main research hypotheses of this work are as follows:

\textbf{Turn-taking and Latency:} Participants will perceive the robot’s turn-taking as natural~\textbf{(H1)}.

\textbf{Back-channeling:} Participants will perceive that the robot’s non-verbal behaviors are effectively synchronized with its conversation~\textbf{(H2a)} and that the robot is actively listening~\textbf{(H2b)}.

\textbf{Adaptive Responses:} Participants will perceive the content of the robot’s responses as adaptive to the flow of the conversation~\textbf{(H3)}.

\textbf{Voice:} Participants will perceive the robot’s voice as appropriate for well-being support~\textbf{(H4a)} and will acknowledge that it adapts to their own vocal tone and emotional expression~\textbf{(H4b)}.

\textbf{Overall Interaction:} Participants will feel comfortable sharing emotions~\textbf{(H5a)} and personal experiences~\textbf{(H5b)} with the robot. They will feel satisfied with the robot’s responses~\textbf{(H5c)}, consider the robot to be empathetic~\textbf{(H5d)}, report feeling positive during the interaction~\textbf{(H5e)}, and feel that the robot helped them appreciate aspects of their lives~\textbf{(H5f)}.

\textbf{Participant Well-being Outcomes:} Interacting with the robot will improve participants’ self-reported levels of gratitude~\textbf{(H6a)} and life satisfaction~\textbf{(H6b)}.

\subsection{Recruitment and Participants}

The study was approved by the Institutional Review Board at the University of Southern California (USC IRB \#UP-25-00176). Inclusion criteria were: being a USC undergraduate, Master’s, or PhD student over 18 years of age, proficient in English, with normal or corrected hearing. Eleven students consented to and participated in the study; 7 self-identified as female and 4 as male, with ages ranging from 18 to 29, except one participant who chose not to disclose their exact age. All participants reviewed the consent information sheet prior to their study session outlining how their data would be collected and used, and consented to participate. They were compensated with a US\$12 Amazon gift card for their time.

\subsection{Procedure}

For each session, before the interaction with the robot began, participants were asked to complete a quantitative pre-test questionnaire (detailed in Section~\ref{sec:questions}), which took approximately five minutes.

Once the interaction began, participants engaged with the robot, which was prompted to guide them through sharing two things they are grateful for and two personal achievements~\cite{axelsson2024robots}. The prompt also instructed the robot to provide positive reinforcement, remain conversational, and offer personalized responses throughout the interaction. This phase lasted approximately 15 minutes.

After the interaction, participants completed a quantitative post-test questionnaire (see Section~\ref{sec:questions}) and then participated in a qualitative, semi-structured interview with a member of the research team to discuss their ratings and overall experience. This final portion took approximately 20 minutes.

\subsection{Quantitative and Qualitative Measures}
\label{sec:questions}
To quantitatively evaluate the robot's usability ~\textbf{(H1–H5}), we designed Likert-scale questions to assess participants’ experiences with the robot, informed by usability constructs and themes identified in prior HRI research~\cite{reimann2024survey,axelsson2024robots}. After the interaction with the robot, participants were asked to rate their experience on a 7-point scale (1 = strongly disagree, 7 = strongly agree) across several dimensions: naturalness of turn-taking~\textbf{(H1)}, synchronization of movements with conversation~\textbf{(H2a)}, active listening~\textbf{(H2b)}, adaptiveness of response content~\textbf{(H3)}, appropriateness of voice for well-being support~\textbf{(H4a)}, adaptiveness of voice tone and emotional expression~\textbf{(H4b)}, comfort in sharing emotions~\textbf{(H5a)} and personal experiences~\textbf{(H5b)}, satisfaction with the robot’s responses~\textbf{(H5c)}, perception of the robot as empathetic~\textbf{(H5d)}, feeling positive during the interaction~\textbf{(H5e)}, and feeling that the robot helped them appreciate aspects of their lives~\textbf{(H5f)}. These questions regarding robot's usability were included as part of the post-test questionnaire.

To quantitatively evaluate~\textbf{H6}, we used Likert-scale questions validated in previous gratitude studies including: 
1) Six-Item Form (GQ-6)~\cite{mccullough2002grateful}; 2) Multi-Component Gratitude Measure (MCGM)~\cite{morgan2017new}; and 3) The Satisfaction With Life Scale (SWLS)~\cite{diener1985satisfaction}. All items from the GQ-6, MCGM, and SWLS used a 7-point scale (1 = strongly disagree, 7 = strongly agree). These questions regarding the participants' self-reported levels of gratitude and life satisfaction were included as part of both the pre-test and post-test questionnaire.

To qualitatively explore participants’ responses to the questionnaires, we conducted a semi-structured interview following the questionnaires. We asked open-ended questions to gather qualitative feedback on potential improvements, including personalization, engagement, and design adjustments. These interviews aimed to identify both the strengths and areas for enhancement in the robot’s interaction capabilities.

\subsection{Data Analysis}

We analyzed robot-related measures \textbf{H1--H5} using both quantitative and qualitative methods. Quantitatively, we used a two-sided one-sample Wilcoxon signed-rank test to assess whether there was a significant difference between the neutral point (4) on the Likert scale and the median of the participants’ responses. Qualitatively, we incorporated insights from the post-interaction semi-structured interviews to contextualize and enrich the quantitative findings.

For participant well-being metrics \textbf{H6}, we conducted a quantitative analysis only. To evaluate significant differences between pre- and post-test scores, we used the two-sided Wilcoxon signed-rank test, a non-parametric method for paired data. This test was selected because it does not assume a normal distribution and is well-suited for evaluating changes in responses collected from the same participants.

For hypotheses that contained multiple statistical tests (i.e., \textbf{H2}, \textbf{H4}, \textbf{H5}, and \textbf{H6}), we applied the Holm–Bonferroni correction using the formula $\alpha' = \frac{\alpha}{m - n + 1}$ to determine the $n^\text{th}$ test's significance within each hypothesis group independently. In all Wilcoxon signed-rank tests, $p$ denotes the two-tailed probability of observing the result under the null hypothesis, and $r$ denotes the effect size (with $r \geq 0.5$ considered large).


\section{Results}
\label{sec:results}
All 11 sessions were completed without major issues. In one session, the system failed to record the participant’s response in the transcript; however, the participant from that session reported that the conversation proceeded smoothly. Two participants reported a single instance of lag during their sessions, likely caused by network issues.

\begin{figure*}[t]
    \centering
    \hspace*{-2.3cm} 
    \includegraphics[width=1.26\textwidth]{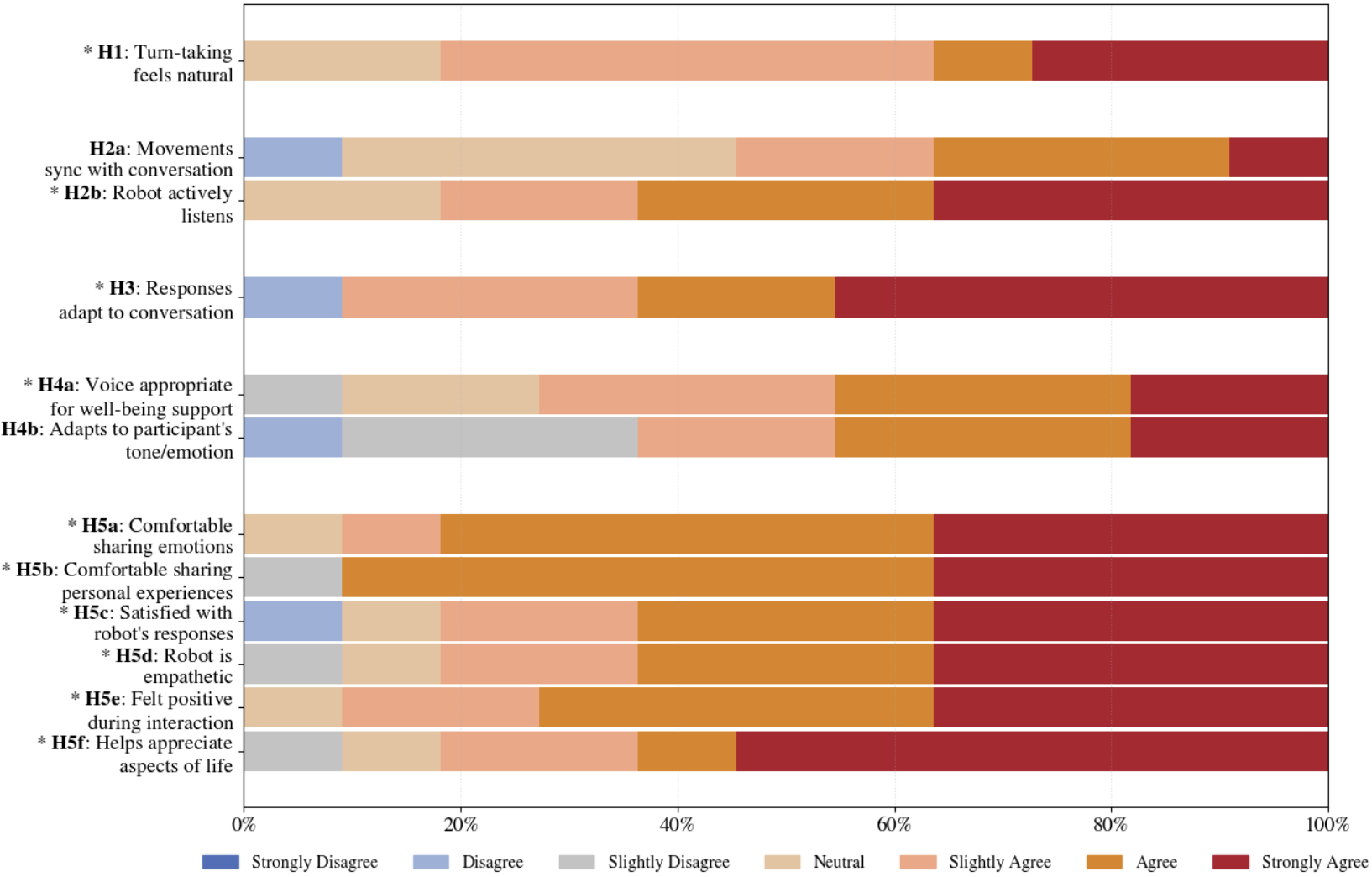}
    \caption{Participant feedback on robot characteristics (* $= p < .05$, ** $= p < .001$)}
    \label{fig:robotfeedback}
\end{figure*}

\textbf{H1 (turn-taking feels natural):} Participants rated turn-taking higher than the neutral midpoint (Mean = 5.45, Median = 5.00). The Wilcoxon signed-rank test revealed a statistically significant difference from the neutral midpoint of 4 ($r$ = 0.870, $p$ = .004).

\textbf{H2 (back-channeling):} The level of movement synchronization with conversation was rated slightly higher than the neutral midpoint (Mean = 4.82, Median = 5.00). For \textbf{H2a}, there was no statistically significant difference from 4 ($r$ = 0.463, $p$ = .125). Participants rated the robot's ability to actively listen higher than the neutral midpoint (Mean = 5.82, Median = 6.00). For \textbf{H2b}, the Wilcoxon test showed a statistically significant difference from 4 ($r$ = 0.870, $p$ = .004).

\textbf{H3 (adaptive responses):} Participants rated the adaptiveness of the robot's responses to conversation higher than the neutral midpoint (Mean = 5.82, Median = 6.00). For \textbf{H3}, there was a statistically significant difference from the midpoint ($r$ = 0.779, $p$ = .010).

\textbf{H4 (voice):} The appropriateness of the robot's voice for well-being support was rated higher than the neutral midpoint (Mean = 5.27, Median = 5.00). For \textbf{H4a}, the difference from 4 was statistically significant ($r$ = 0.704, $p$ = .020). The robot's ability to adapt to the participant's tone and emotion was rated slightly higher than the neutral midpoint (Mean = 4.82, Median = 5.00). For \textbf{H4b}, the Wilcoxon test showed no statistically significant difference from 4 ($r$ = 0.414, $p$ = .170).

\textbf{H5 (overall interaction):} Participants rated their comfort sharing emotions with the robot higher than the neutral midpoint (Mean = 6.09, Median = 6.00). For \textbf{H5a}, the difference from neutral was statistically significant ($r$ = 0.934, $p$ = .002). Participants rated their comfort sharing personal experiences higher than the neutral midpoint (Mean = 6.09, Median = 6.00). For \textbf{H5b}, the test found a statistically significant difference ($r$ = 0.934, $p$ = .002). Satisfaction with the robot’s responses was rated higher than the neutral midpoint (Mean = 5.64, Median = 6.00). For \textbf{H5c}, the difference was statistically significant ($r$ = 0.729, $p$ = .016). The robot’s empathy was rated higher than the neutral midpoint (Mean = 5.73, Median = 6.00). For \textbf{H5d}, the Wilcoxon test indicated a statistically significant difference ($r$ = 0.802, $p$ = .008). Participants reported feeling more positive during the interaction compared to the neutral midpoint (Mean = 6.00, Median = 6.00). For \textbf{H5e}, participants rated feeling positive during the interaction higher than the neutral midpoint (Mean = 6.00, Median = 6.00), with a statistically significant difference ($r$ = 0.934, $p$ = .002). For \textbf{H5f}, participants rated the robot as helping them appreciate aspects of their lives higher than the neutral midpoint (Mean = 6.00, Median = 7.00), with a statistically significant difference ($r$ = 0.802, $p$ = .008).


\textbf{H6 (participant well-being outcomes):} Participants completed well-being surveys before and after the interaction. All measures showed statistically significant improvements post-intervention. 
For \textbf{H6a}, general gratitude increased from 5.49 to 5.86 (Median = 6.00), showing a statistically significant improvement ($r$ = -0.590, $p$ = .027). Life satisfaction improved from 4.24 to 4.86 (Median = 4.80). For \textbf{H6b}, this difference was statistically significant ($r$ = -0.489, $p$ = .022).

\section{Discussion}
This work presents a first small-scale exploration of integrating end-to-end speech-language models (SLMs) with socially assistive robots (SARs) to support well-being. 

{\it Hypothesis testing results.} Our results show that turn-taking was rated significantly above neutral (\textbf{H1} is supported). Although participants preferred the robot’s movement over static behavior, its synchronization with conversation was rated no differently than neutral (\textbf{H2a} is not supported). In contrast, participants rated the robot’s active listening ability as significantly above neutral (\textbf{H2b} is supported). Participants considered the robot’s responses as adaptive (\textbf{H3} is supported), though interview data suggested room for improvement in personalization. The robot's voice suitability for well-being support was rated significantly above neutral (\textbf{H4a} is supported), while its adaptation to participants’ vocal tone and emotional expression was not observed (\textbf{H4b} is not supported). Participants reported high comfort in sharing emotions (\textbf{H5a} is supported) and personal experiences (\textbf{H5b} is supported), were satisfied with the robot’s responses (\textbf{H5c} is supported), viewed the robot as empathetic (\textbf{H5d} is supported), reported feeling positive during the interaction (\textbf{H5e} is supported), and felt that the robot helped them appreciate aspects of their lives (\textbf{H5f} is supported). Finally, all well-being outcomes showed statistically significant improvements from pre- to post-test, including general gratitude (\textbf{H6a} is supported) and life satisfaction (\textbf{H6b} is supported).

\textit{End-to-end SLMs can effectively support turn-taking in real-time SAR interactions.} As reported in Section~\ref{sec:results}, for~\textbf{H1}, our quantitative results show that participants perceived the robot’s turn-taking as natural. These findings are further supported by our qualitative analysis of interview transcripts, where no participants reported issues with turn-taking or latency. Instead, interactions were frequently described as natural and fluid. One participant remarked, “It was as similar as if I was speaking to a person.” These results suggest that integrating an end-to-end SLM in place of the cascaded dialogue pipeline may be a promising approach for enabling more effective real-time turn-taking in SARs.

\textit{Back-channeling robot movements are helpful, but more diverse and synchronized generation is needed to reduce repetitiveness.} For \textbf{H2a} and \textbf{H2b}, our results show that participants perceived the robot as actively listening during the interaction but did not feel that its movements were well-synchronized with the conversation. Our qualitative findings revealed that participants preferred movement over a static posture; however, repetitive nodding that lacked synchronization with the conversational context was often perceived as robotic. As one participant noted, “There was one part where it was nodding a lot, and then I felt it looked too much like a robot.” Some participants also mentioned that motor noise during pauses was distracting. While back-channeling movements were generally seen as beneficial, repetitive or unrefined behaviors may have detracted from the overall experience. These findings suggest a need for generating more diverse robot movements synchronized with both the robot’s and the user's speech.

\textit{The content of robot responses adapts to participant input, albeit in a rigid, generic, and predictable structure.} For \textbf{H3}, our quantitative results showed that participants perceived the robot’s responses as adaptive to the conversation. However, our qualitative results revealed that while participants agreed the responses were relevant, they also described them as “too structured.” One participant remarked, “I felt like it heard me, but I didn’t feel understood.” Suggested improvements from the participants included more selective summarization and varied phrasing. This suggests that a more sophisticated prompting framework or fine-tuning of the models may be needed to better align SLM responses with best practices from mental health experts. 


\textit{The robot’s voice is perceived as well-suited for well-being support.} For \textbf{H4a}, our results found that participants perceived it as appropriate for the context of well-being support. However, some noted a mismatch between the voice and the robot’s childlike appearance. Suggested improvements included voice customization and the addition of natural interjections (e.g., “hmm,” “oh”) to enhance authenticity. As one participant remarked, “I just wish the voice sounded more like a friend than a sort of counselor.”

\textit{The tone and emotion of the voice did not appear to adapt dynamically to participants.} For \textbf{H4b}, participants reported that the SLM’s voice output did not adapt its tone or emotional expression to the context of the conversation. This likely reflects limitations in the underlying SLM, which still lacks the expressiveness and capability to dynamically adjust tone based on conversational cues. Participants described the robot’s voice as flat and lacking emotional nuance, with one noting, “I didn’t really see a big change in the tone.” They suggested that more accurate mirroring of participants’ mood and energy levels could enhance engagement.

\textit{Overall, participants found the interaction with the robot to be comfortable, emotionally supportive, and positively engaging, though sometimes limited in depth.} For \textbf{H5a} and \textbf{H5b}, participants reported that they felt comfortable in sharing both emotions and personal experiences with the robot. Participants shared that speaking with a robot made them feel more at ease opening up emotionally. As one participant explained, “I felt comfortable because it’s not a real person—it can’t judge you.” Many also appreciated the robot’s consistent positivity. For \textbf{H5c} and \textbf{H5d}, participants were satisfied with the robot’s responses and perceived it as empathetic; however, our qualitative results suggested this perception was largely influenced by modest expectations. Participants did not anticipate human-level dialogue and were therefore more easily satisfied. For \textbf{H5e}, participants perceived the robot as positive throughout the interaction. Although the research team was initially concerned about the risk of model hallucinations or inappropriate responses, no such instances occurred during the study. In fact, several participants noted that the robot’s positivity occasionally felt excessive. Finally, for \textbf{H5f}, participants agreed that the robot helped them reflect on and appreciate aspects of their lives. This outcome aligns with the goals of the well-being intervention and supports prior findings, providing preliminary evidence for the potential of integrating end-to-end SLMs into SARs.


\textit{A 15-minute interaction helped improve short-term well-being outcomes.} As shown in Section~\ref{sec:results}, for \textbf{H6}, both self-reported ratings of general gratitude and life satisfaction significantly improved from pre-test to post-test. This was somewhat unexpected given the brief 15-minute interaction, and may reflect a novelty effect from the robot. Additionally, we acknowledge that expecting substantial changes in life satisfaction following a single 40-minute session is not realistic. Accordingly, these findings should be interpreted as preliminary indicators of potential for long-term impact and explored further. Nonetheless, they suggest the promise of speech-based SARs to positively support well-being. As one participant noted, “It helped me reflect… and made me feel grateful.”

\subsection{Limitations}
We acknowledge several limitations of this work. First, this study is a small-scale preliminary exploration (N = 11) aimed at validating the potential of end-to-end speech-language models (SLMs) to enable more effective real-time dialogue for socially assistive robots (SARs). Second, the user study followed a single-session design, with each session lasting a maximum of 15 minutes and no follow-up interactions. As a result, the positive outcomes observed for \textbf{H6} may have been influenced by a novelty effect. To confirm the long-term impact of SLM-enabled SARs, larger-scale, longitudinal, and ecologically valid field studies are needed. In addition, we did not include a baseline condition to directly compare end-to-end SLMs with a cascaded dialogue pipeline. Although our results show promising potential, future work should incorporate such comparisons to validate the observed improvements. 

While turn-taking was rated positively, participants were required to press and hold a mouse button to speak and release it to yield their turn. This design simplified the turn-taking process but does not reflect the challenges of free-form conversational turn-taking. Future research should further investigate SLMs’ capabilities in managing natural, real-time conversational flow with free-form turn-taking. Moreover, the robot was limited to fixed movements; future work should enhance non-verbal synchrony by incorporating multi-modal signal alignment and gesture-generation models. Lastly, our implementation relied solely on basic prompt engineering, without the use of a sophisticated prompting framework. Further exploration is needed to evaluate how more advanced prompting or fine-tuning techniques might enhance SLM performance in SAR contexts.


\section{Conclusion}
This paper presented an exploration of integrating real-time, end-to-end speech-language models (SLMs) with socially assistive robots (SARs), providing preliminary results and insights from a small-scale user study that can inform future development of SLM-enabled SARs. While our findings highlight the promising potential of SLMs to support more empathetic feedback, natural turn-taking, back-channeling, and adaptive responses in SARs, further work is needed to enable more synchronized real-time movement, align SLM outputs with best practices from mental health experts, and improve expressive, adaptive voice generation.

\subsubsection*{Acknowledgments.}
\small
This work is supported in part by the National Science Foundation (IIS-1925083), departmental funding from the University of Southern California, and the Center for Undergraduate Research in Viterbi Engineering (CURVE) Fellowship at the University of Southern California.

\bibliographystyle{splncs04}
\bibliography{root}

\end{document}